\documentclass[runningheads]{llncs}

\usepackage{eccv}

\usepackage{eccvabbrv}

\usepackage{graphicx}
\usepackage{booktabs}
\usepackage{amsmath,amssymb}

\usepackage{tikz}
\usepackage{comment}
\usepackage{amsmath,amssymb} 
\usepackage{color}
\usepackage{booktabs}
\usepackage{mathtools}
\usepackage{amsmath}
\usepackage{relsize}
\usepackage{multirow}

\usepackage{makecell}

\usepackage{adjustbox}
\usepackage{graphicx}
\usepackage{fix-cm}
\makeatletter
\def\thickhline{%
  \noalign{\ifnum0=`}\fi\hrule \@height \thickarrayrulewidth \futurelet
   \reserved@a\@xthickhline}
\def\@xthickhline{\ifx\reserved@a\thickhline
               \vskip\doublerulesep
               \vskip-\thickarrayrulewidth
             \fi
      \ifnum0=`{\fi}}
\makeatother
\newlength{\thickarrayrulewidth}
\usepackage{setspace}
\usepackage{bm}

\usepackage{eccvabbrv}

\usepackage{graphicx}
\usepackage{booktabs}
\usepackage{multirow,multicol} 
\usepackage{algorithm} 
\usepackage{algpseudocode} 
\usepackage{xcolor,colortbl} 

\usepackage{color}
\usepackage{highlight}
\usepackage{fix-cm}
\makeatletter
\def\thickhline{%
  \noalign{\ifnum0=`}\fi\hrule \@height \thickarrayrulewidth \futurelet
   \reserved@a\@xthickhline}
\def\@xthickhline{\ifx\reserved@a\thickhline
               \vskip\doublerulesep
               \vskip-\thickarrayrulewidth
             \fi
      \ifnum0=`{\fi}}
\makeatother
\definecolor{darkblue}{rgb}{0, 0, 0.7}

\usepackage{makecell}

\usepackage{adjustbox}
\usepackage{graphicx}
\usepackage{fix-cm}
\makeatletter
\def\thickhline{%
  \noalign{\ifnum0=`}\fi\hrule \@height \thickarrayrulewidth \futurelet
   \reserved@a\@xthickhline}
\def\@xthickhline{\ifx\reserved@a\thickhline
               \vskip\doublerulesep
               \vskip-\thickarrayrulewidth
             \fi
      \ifnum0=`{\fi}}
\makeatother

\usepackage{tikz}
\newcommand{\tikzxmark}{%
\tikz[scale=0.23] {
    \draw[line width=0.7,line cap=round] (0,0) to [bend left=6] (1,1);
    \draw[line width=0.7,line cap=round] (0.2,0.95) to [bend right=3] (0.8,0.05);
}}
\newcommand{\tikzcmark}{%
\tikz[scale=0.23] {
    \draw[line width=0.7,line cap=round] (0.25,0) to [bend left=10] (1,1);
    \draw[line width=0.8,line cap=round] (0,0.35) to [bend right=1] (0.23,0);
}}

\usepackage[accsupp]{axessibility}  

\usepackage[pagebackref,breaklinks,colorlinks,citecolor=eccvblue]{hyperref}

\usepackage{orcidlink}

\usepackage[accsupp]{axessibility}  

\usepackage{hyperref}

\usepackage{orcidlink}

\begin{document}


\title{Temporal Event Stereo via Joint Learning with Stereoscopic Flow}

\newcommand\CoAuthorMark{\footnotemark[\arabic{footnote}]}
\author{Hoonhee Cho\orcidlink{0000-0003-0896-6793}\thanks{Equal contribution.} \and
Jae-Young Kang\orcidlink{0009-0002-9537-3813}\protect\CoAuthorMark \and
Kuk-Jin Yoon\orcidlink{0000-0002-1634-2756}}

\authorrunning{Cho et al.}

\institute{Korea Advanced Institute of Science and Technology\\
\email{\{gnsgnsgml, mickeykang, kjyoon\}@kaist.ac.kr}\\
}

\maketitle

\begin{abstract}
Event cameras are dynamic vision sensors inspired by the biological retina, characterized by their high dynamic range, high temporal resolution, and low power consumption. These features make them capable of perceiving 3D environments even in extreme conditions. Event data is continuous across the time dimension, which allows a detailed description of each pixel's movements. To fully utilize the temporally dense and continuous nature of event cameras, we propose a novel temporal event stereo, a framework that continuously uses information from previous time steps. This is accomplished through the simultaneous training of an event stereo matching network alongside stereoscopic flow, a new concept that captures all pixel movements from stereo cameras. Since obtaining ground truth for optical flow during training is challenging, we propose a method that uses only disparity maps to train the stereoscopic flow. The performance of event-based stereo matching is enhanced by temporally aggregating information using the flows. We have achieved state-of-the-art performance on the MVSEC and the DSEC datasets. The method is computationally efficient, as it stacks previous information in a cascading manner. The code is available at \url{https://github.com/mickeykang16/TemporalEventStereo}.

\end{abstract}

\section{Introduction}
\label{sec:intro}

Depth estimation~\cite{Laga2020ASO,Scharstein2004ATA} is considered a fundamental problem within the field of computer vision, which can be adapted to a wide range of practical applications, including robotics and autonomous driving. Stereo matching~\cite{Chang2018PyramidSM,Xu2020AANetAA,Zhang2019GANetGA,Kendall2017End,Guo2019GroupWiseCS,Cheng2020HierarchicalNA,zhang2023temporalstereo, xu2020aanet, weinzaepfel2023croco, xu2023iterative, li2022practical}, a prominent technique in depth estimation, identifies correspondences between rectified stereo images to compute depth. Recent stereo matching algorithms are tailored to conventional frame-based cameras. Yet, these sensors have limitations including high power consumption, limited dynamic range, and low data rate, constraining their use in mobile devices and edge computing platforms. On the other hand, drawing inspiration from the biological retina, the recently emerged dynamical vision sensor, also known as an event camera~\cite{Brandli2014A2,Lichtsteiner2008A11}, addresses these issues by transmitting events representing instantaneous changes in pixel intensity with millisecond-level temporal resolution. Event cameras can avoid motion blur caused by rapid object movement, and their ability to operate with high dynamic range and low power consumption could make them a game changer in autonomous driving and mobility systems.

By the nature of event cameras, transmitted events are sparse in space while containing dense temporal information. To accommodate the sparsity of the spatial domain, existing methods actively utilize the time information of events in the task of stereo matching. 
For example, Se-CFF~\cite{nam2022stereo}  proposed a method that stacks events from a small to a large number and concentrates them into one, through an attention-based network. Another example, DTC~\cite{zhang2022discrete} adopts a recurrent structure~\cite{funahashi1993approximation} to encode events sequentially, enriching current features.
Although they have improved performance from temporal aggregation approaches, they lack a specialized approach to stereo matching. Specifically, temporal information transfer is done independently in left and right event, and limited to low-level representations (\eg,~feature map). 
Therefore, the limited data transfer results in sub-optimal performance.
We made a breakthrough by proposing temporal event stereo, which uses a flow-based temporal aggregation.

\begin{figure*}[t]
\begin{center}
\includegraphics[width=.82\linewidth]{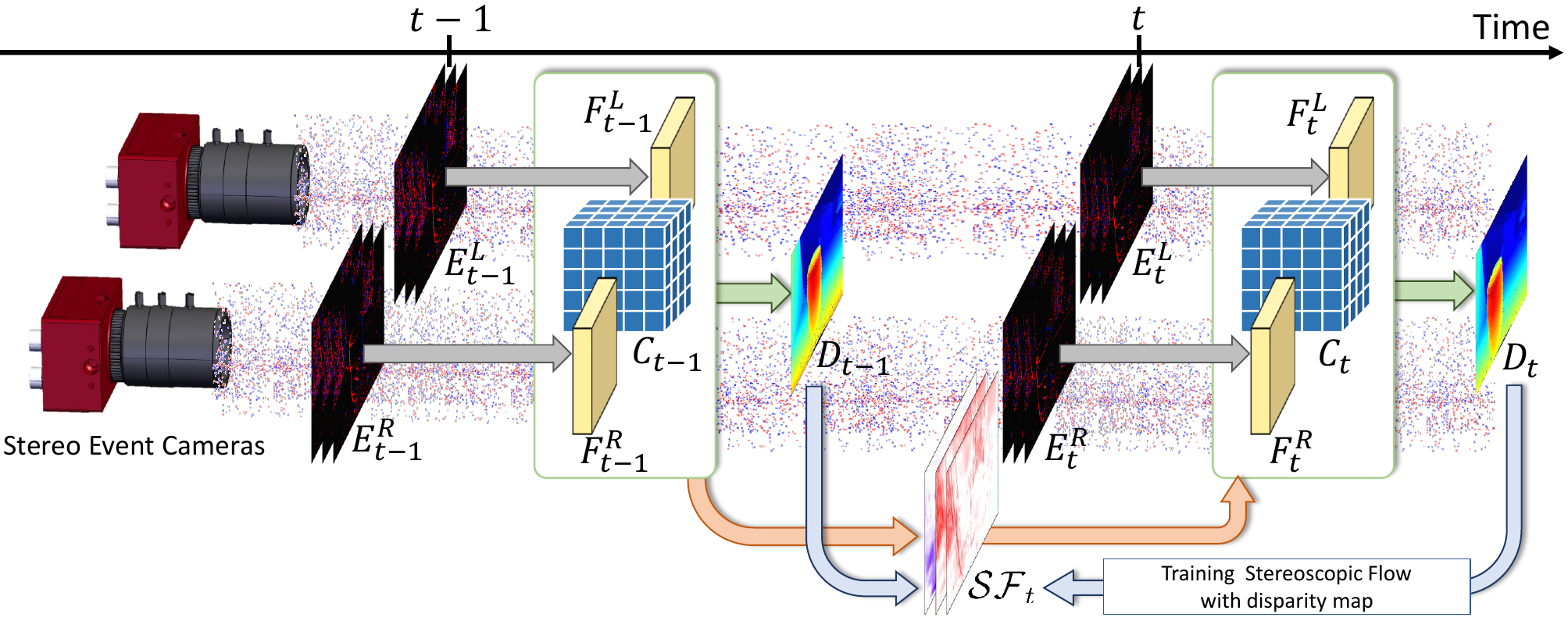}
\caption{\textbf{Overview of the temporal event stereo.} We can accurately and efficiently estimate dense disparity by propagating previously computed information to the present through stereoscopic flow ($\mathcal{SF}$).}
\label{fig:teaser}
\end{center}
\end{figure*}

Unlike the typical optical flow~\cite{hur2019iterative,ilg2017flownet,jiang2021learning,sun2018pwc,teed2020raft,weinzaepfel2013deepflow,liu2018adaptive, zhu2019unsupervised,bardow2016simultaneous, gehrig2021raft} that calculates pixel displacement of a single camera, we define ``stereoscopic flow'' considering the specificity of stereo matching. Each corresponding point from two stereo cameras lies on a single horizontal epipolar line. Even if there is vertical pixel displacement over consecutive time steps, the matching points are placed on an epipolar line.
Stereoscopic flow involves a triplet flow, defined in rectified stereo image pairs, consisting of two independent horizontal pixel displacements for each camera and a single vertical pixel displacement. As shown in Fig.~\ref{fig:teaser}, we reuse previous features and cost volumes by temporally propagating them through stereoscopic flow.
The previous model~\cite{zhang2022discrete}, which uses recurrent structure, requires independent networks to aggregate different types of representations. However, our approach enables the aggregation of not only low-level features but also stereo-dependent information such as a cost volume with minimal effort. Instead of requiring independent modules for each component, we can efficiently retrieve previous information through a warping operation using the once-calculated stereoscopic flow.
Furthermore, without using additional and independent two optical flow networks, we can calculate temporal pixel movement with just a single stereoscopic flow network composed of very few convolutions as we jointly learn using dense representation features learned from the stereo network.
To ensure the stable training of stereoscopic flow, we introduce a novel and effective temporal disparity consistency loss. This loss facilitates the training of stereoscopic flow when two stereo cameras and a disparity map are available, offering a strategy to train stereoscopic flow in the absence of optical flow ground truth.
Finally, we propose feature warping and cost volume warping to merge previous information with current inputs effectively, achieving state-of-the-art (SOTA) performance on the MVSEC and DSEC datasets.

Our contributions are summarized as follows:
1) We propose feature warping and cost volume warping to fuse previous information through stereoscopic flow effectively.
2) We propose a framework that jointly trains stereoscopic flow, as defined by a pair of stereo cameras, and stereo matching.
3) We introduce a temporal disparity consistency loss that enables the learning of stereoscopic flow even without optical flow ground truth.
4) Our approach significantly improves performance while being computationally efficient and achieving state-of-the-art (SOTA) on MVSEC and DSEC benchmarks.

\section{Related Works}
\noindent
\textbf{Event-based Stereo Matching.}
Unlike traditional cameras, event cameras~\cite{gallego2020event} capture the scene as a stream of sparse events, representing per-pixel intensity changes exceeding a set threshold. Their asynchronous nature makes them less susceptible to motion blur~\cite{lin2020learning, sun2022event, sun2023event, Xu_2021_ICCV,zhang2021fine, zhang2022unifying, pan2019bringing, kim2024frequency} from rapid scene changes or movements. Event cameras, with their enhanced dynamic range~\cite{rebecq2019high, han2020neuromorphic,  perot2020learning}, capture scenes that traditional cameras may miss, such as night scenes~\cite{xia2023cmda, zhang2020learning, alonso2019ev, wang2021dual, cho2022selection}, making them ideal for driving scenarios.

Early event-based stereo matching utilized the hand-crafted method \cite{kogler2011event,camunas2014use,zou2016context,zou2017robust,piatkowska2013asynchronous,rogister2011asynchronous,zhu2018realtime,piatkowska2017improved,eomvs} to determine corresponding events between asynchronous streams. Recent works \cite{nam2022stereo, ahmed2021deep, zhang2022discrete, cho2023learning} adopt deep learning-based approaches by processing events into image-like forms \cite{liu2018adaptive, zihao2018unsupervised, manderscheid2019speed, lagorce2016hots, maqueda2018event, cho2022event, mostafavi2021learning, nguyen2019real, zhu2019unsupervised, wang2019event, choi2020learning, mostafavi2021e2sri, cho2022selection} to accommodate the property of the sparsity of event data and to benefit from the knowledge of image-based deep learning stereo methods.
To overcome the shortcomings of an event density imbalance across spatial dimensions, Se-CFF~\cite{nam2022stereo} proposed the multi-density stack event representation. 
Furthermore, DTC~\cite{zhang2022discrete} aggregated sequential event information at the feature level using the discrete time convolutional module. 
As they have focused on low-level feature implementation, there is still potential for further advancement. It is possible to expand existing methods~\cite{nam2022stereo, zhang2022discrete} so that they can deal with higher-level representations (\eg cost volume) by adding aggregation modules. However, the more information crosses from the past, the more computational cost should be added to the model, which forces a trade-off between information and complexity. 
To make a breakthrough, we propose a method of flow-based temporal aggregation, a cost-effective method to gather multiple types of information.
Our flow-based aggregation shares a one-time estimated flow in both feature warping and cost volume warping, adding minimal computation.

\noindent
\textbf{Learning with Event-based Optical Flow.}
Event cameras efficiently capture per-pixel changes despite their sparse data, proving highly useful for optical flow estimation~\cite{zhu2018ev, pan2020single, wan2022learning, ding2022spatio,  almatrafi2020distance,   benosman2013event, gehrig2024dense, hu2022optical, paredes2023taming, liu2023tma, Luo_2023_ICCV, Wan_2023_ICCV}. Their capability extends to capturing motion effortlessly, aiding in tasks like video frame interpolation~\cite{Kim_2023_CVPR, tulyakov2021time, tulyakov2022time, cho2024tta}, image reconstruction~\cite{paredes2020back, fox2024unsupervised, cho2023label}, and motion deblurring~\cite{jiang2020learning, cho2023non, xu2021motion}. These works focus on a main task while setting optical flow from events as an auxiliary task for joint learning to assist. The optical flow is used for temporal propagation and alignment, boosting the performance of the main task.

In our work, stereoscopic flow is estimated from stereo events to determine the pixel movement of stereo cameras. Stereoscopic flow is used to temporally transfer the features and cost volume to the next time step. Previous works~\cite{paredes2020back, xu2021motion} designed independent flow estimation networks. However, inspired by the fact that both stereo and flow networks match features between frames, our flow network shares features from the stereo network, allowing for cost-effective design.
To the best of our knowledge, our method is the first to propose the joint learning of event-based optical flow for stereo matching.

\noindent
\textbf{Event-based Optical Flow without Ground-truth.} 
Events are continuous points containing temporal information, and by analyzing their motion over time, the displacement of each pixel can be effectively calculated.
Some studies~\cite{gallego2018unifying, zihao2018unsupervised, ye2018unsupervised, shiba2022secrets, paredes2019unsupervised} have demonstrated distinction in optical flow through unsupervised learning by effectively utilizing the properties of events.
Loss functions that solely rely on events face convergence issues depending on the initialization, as mentioned in ~\cite{nam2022stereo, shiba2022secrets, Shiba2022EventCI}. This problem occurs when trained in an end-to-end manner with a stereo matching network. Also, as it is an edge-centric loss, we propose a new temporal disparity consistency loss.

\section{Methodology}
\subsection{Preliminary}
Event streams of left and right cameras between $t-1$ and $t$ are represented as $\mathcal{E}_t^L$ and $\mathcal{E}_t^R$, respectively. An event stream is a sequence of four-dimensional vectors $ (i, \ j, \ p, \ \tau) $. $(i, \ j)$ represents pixel coordinate while $p \in \{-1, 1\}$ and $\tau \in (t-1, t]$ represents polarity and time respectively. \\
\textbf{Event Representation.} Following~\cite{zhu2019unsupervised}, event streams $\mathcal{E}_t=\{(i, \ j, \ p, \ \tau)|$$t-1\leq\tau<t\}$ are stacked to generate an event voxel grid $E_t(b, i, j) \in \mathbb{R}^{B \times H \times W}$ with discretized time dimension B, scaled to $b\in[0, B-1]$.\\
\noindent
\textbf{Stereoscopic Flow.}\label{sec:stereo_flow} Given a pair of rectified stereo event voxels  $(E_t^L, E_t^R)$,
matching points in stereo voxels are on an epipolar line, which is a single horizontal line for rectified stereo. Since the epipolar constraint remains for all stereo pairs, it can be assumed that each matching pixels in left and right voxels experience the same vertical shift, while their horizontal shift may differ due to depth changes. Therefore, the flow of stereo voxel pairs can be expressed as a set of two horizontal flows for each voxel and a single shared vertical flow. We use the notation of 
$
\mathcal{SF}_t=\{
\underset{t\rightarrow t-1}{\Delta x^L},
\underset{t\rightarrow t-1}{\Delta x^R},
\underset{t\rightarrow t-1}{\Delta y}\} \in \mathbb{R}^{3 \times H \times W},
$
for our stereoscopic flow. Please note that all flows are defined as backward flows, from time frame $t$ to $t-1$, to utilize backward warping. The terms stereoscopic flow, stereo flow, and $\mathcal{SF}$ will be used interchangeably throughout the paper.

\begin{figure*}[t]
\begin{center}
\includegraphics[width=0.99\linewidth]{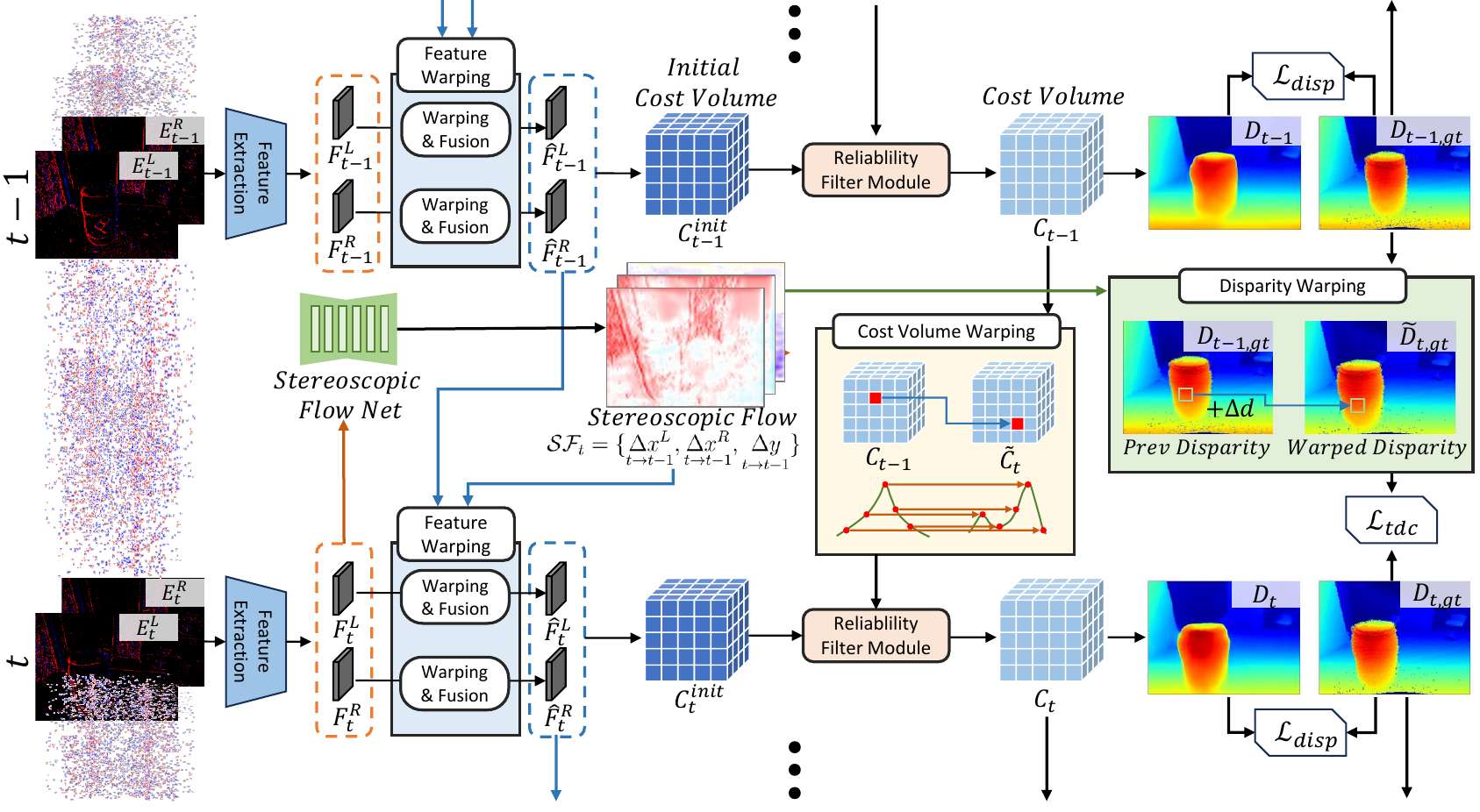}
\caption{\textbf{Architecture overview of the proposed temporal event stereo.} In each time step, information from the preceding moment is warped via stereoscopic flow and fused with the current information, boosting the intermediate representations such as feature map and cost volume. Stereoscopic flow training does not require the ground truth optical flow but receives a supervision signal from ground truth disparity.}
\label{fig:acrhitecture}
\end{center}
\end{figure*}

\subsection{Overall Architecture Design}
As shown in Fig.~\ref{fig:acrhitecture}, we design the joint framework for the temporal event-based stereo and flow network, that can be trained in an end-to-end manner. Following the well-working art~\cite{psmnet, Guo2019GroupWiseCS, acvnet, Xu2020AANetAA, Zhang2019ga}, the main streamline of the stereo network is composed of four commonly used parts: 1) feature extraction, 2) cost volume generation, 3) cost aggregation, and 4) cost refinement. Conventional stereo modules are omitted in Fig.~\ref{fig:acrhitecture} for brevity and ease of understanding.
\\
\textbf{Feature Extraction. } At each time step, stereo event voxel pair $(E_t^L$, $E_t^R)$ is fed into the weight-shared feature encoder. These initial features $(F_t^L, F_t^R)$ are concatenated and used to predict stereoscopic flow. With stereoscopic flow, the previous event features are warped and fused with the current ones. 
\\
\textbf{Stereoscopic Flow Network.} Stereoscopic flow, the key feature of our design, plays an important role when aggregating various representations from the previous time step. Unlike the RNN base temporal aggregation methods from~\cite{zhang2022discrete} that require extra RNN blocks for different representations, flow can be used to warp various types (\eg,~feature, cost volume, reliability map, and disparity) with minimal extra computational cost. 
The stereoscopic flow network shares encoded event features from the stereo network, allowing for using high representation features in feature matching. We can estimate stereo flow with a single network consisting of only 9 layers of $3\times3$ 2D convolutions, enjoying the temporally dense property of event data. We concatenate \(F_t^L\) and \(F_t^R\) and feed them into the stereoscopic flow network to obtain the stereo flow. 

The stereoscopic flow is trained with a proposed temporal disparity consistency loss (see Eq.(\ref{equ:loss_tdc})) of ground-truth (GT) disparity. 
Temporal disparity consistency loss is a special flow loss that we introduce to train the stereoscopic flow and disparity collaboratively without using auxiliary GT optical flow. The error between previous GT disparity, warped from previous time $t-1$ to current time $t$, and current GT disparity defines the disparity consistency loss. 
\\
\textbf{Cost Volume Construction and Aggregation. } Following~\cite{Kendall2017End}, a pair of temporally fused features is channel-wise concatenated across all disparity levels, resulting in an initial cost volume $C_t^{init} \in \mathbb{R}^{C \times D \times H \times W} $. It passes two cost aggreagation modules which are lightweight versions of the 3D hourglass~\cite{Guo2019GroupWiseCS}. Initial disparities, $D_t^0$ and $D_t^1$ are predicted after each module. Note that $D_t^0$ and $D_t^1$ are generated for supervision only during the training process for robust cost volume generation and are not computed during inference. 
\\
\textbf{Cost Refinement via Temporal Aggregation.} 
The cost volume is a key representation module that holds task-specific 3D information crucial for stereo matching. Therefore, we propose a way to benefit from the past cost volume. The 4D past cost is warped to the current time frame with our novel stereoscopic flow. Then, the entropy-based reliability fusion module efficiently fuses current and past costs. An hourglass-like refinement block finally refines the fused cost, making a precise disparity prediction.\\
\textbf{Disparity Regression. } A channel aggregation block squeezes the channel of the fused cost volume, resulting in a probability volume. Finally, the predicted disparity is computed by the soft argmin function~\cite{Kendall2017End}.

\subsection{Benefit from the Past}
\label{sec:warping}
Warping and fusion of representations from the previous time steps enrich the intermediate representations (\eg feature, cost volume), improving stereo matching performance cost-effectively. In our design, we propose two modules for temporal information warping: feature warping and cost volume warping. Intermediate representations of $t-1$ are warped to current time $t$ with stereoscopic flow.\\
\textbf{Feature Warping.} 
Events are triggered by changes in intensity, which are mostly generated by ego-motion. Often, due to little motion or static scenes, the current scene may receive too few events. This can be compensated by temporally accumulating information from the past. At each time step, previous features are warped in spatial $(H, W)$ dimensions, keeping the channel value of each pixel. We can obtain the warped feature $\hat{F}_t^L$ by warping past feature using the backward flow $f_t^L$ and warping function $\mathcal{W}_s$, as follows: \(\hat{F}_t^L=\mathcal{W}_s(F_{t-1}^L,\,f_t^L)\).
The spatial warping function $\mathcal{W}_s$, which is a typical backward warping, with backward flow components $f_t^L=\{\underset{t\rightarrow t-1}{\Delta x^L}, \underset{t\rightarrow t-1}{\Delta y}\}$ from $\mathcal{SF}_t$ can be defined as:
\begin{equation}
\begin{aligned}
\mathcal{W}_s(F_{t-1}^L,\,f_t^L)(c,\,h,\,w)
=F_{t-1}^L(c,\ h +\underset{t\rightarrow t-1}{\Delta y}(h,\,w), \ w+\underset{t\rightarrow t-1}{\Delta x^L}(h,\,w)),
\label{equ:spatial_warp}
\end{aligned}
\end{equation}
where $F_{t-1}^L, \hat{F}_{t}^L \in \mathbb{R}^{C \times H \times W}$. The same function $\mathcal{W}_s$ is applied to the previous right feature $F_{t-1}^R$ with flow $f_t^R=\{\underset{t\rightarrow t-1}{\Delta x^R}, \underset{t\rightarrow t-1}{\Delta y^R}\}$. 
$\underset{t\rightarrow t-1}{\Delta y^R} \in \mathbb{R}^{H\times W}$ is a vertical flow of the right frame, estimated along with $\mathcal{SF}_t$ for feature warping.
Warped features, $\hat{F}_t^L$ and $\hat{F}_t^R$, are concatenated in channel dimension with the current event features $F_t^L$ and $F_t^R$, respectively. Then, $3 \times 3$ 2D convolution is applied to squeeze the channel to its original size.\\
\noindent
\textbf{Cost Volume Warping.} 
Cost volume is a core structure that encloses the information on the matching similarity of a stereo pair.
Hence, cost volume summarizes 3D information about the scene and undergoes extensive computations to achieve an optimal aggregation. Many existing works~\cite{acvnet, nam2022stereo, zhang2022discrete} discard the cost volume after computing it for a static time, even if most of the 3D depiction of a scene remains in future frames. We continuously reuse and temporally aggregate cost volumes to achieve a cost-effective design.

\begin{figure*}[t]
\begin{center}
\includegraphics[width=0.88\linewidth]{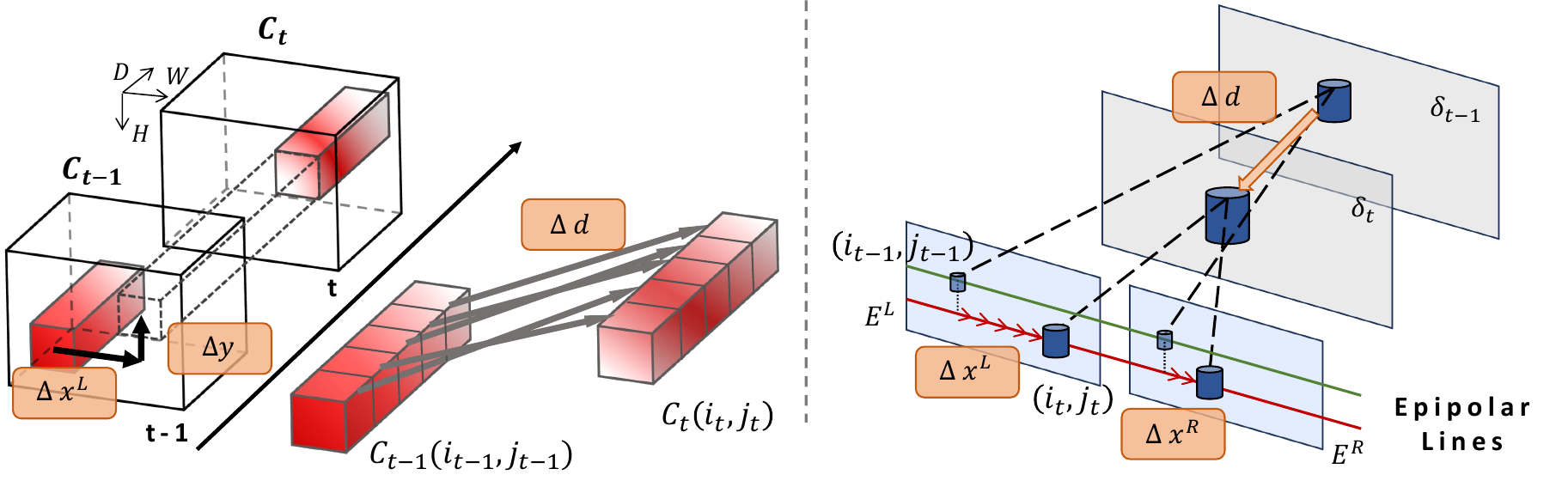}
\caption{\textbf{Cost Volume Warping. } Cost volume warping with 3-dimensional flow $\{\Delta d, \Delta x^L, \Delta y\}$ (left). Relation between disparity flow and stereoscopic flow (right).}
\label{fig:cost_warp}
\end{center}
\end{figure*}

The previous cost volume $C_{t-1} \in \mathbb{R}^{C \times D \times H \times W}$ cannot be directly reused without warping, as it is misaligned to the current initial cost volume $C_t^{init}$. 
Two cost volumes in different time steps are misaligned in two ways. First, as the relative pose between an object and cameras differs, there should be spatial location misalignments on two cost volumes. This spatial movement can be easily estimated as $\{\underset{t\rightarrow t-1}{\Delta x^L}, \underset{t\rightarrow t-1}{\Delta y}\}$ since a cost volume is defined on the left frame coordinate similar to the feature warping process. Second, as the object distance changes, the D-dimension of cost volume suffers a shift. Assume a 3D coordinate $(\delta_{t-1}, i_{t-1}, j_{t-1})$ of a previous cost volume $C_{t-1}$ where $\delta_{t-1} \in [1, D], i_{t-1} \in [1,  H], j_{t-1} \in [1, W]$. A coordinate $(\delta_{t-1}, i_{t-1}, j_{t-1})$ refers to the matching score of stereo pairs for a disparity candidate $\delta_{t-1}$. Therefore, the disparity change of an object would redistribute the matching scores across all disparity candidates on $D$-dimension.

To compensate misalignments on cost volumes, we propose a 3D flow $f_t^C=\{\underset{t\rightarrow t-1}{\Delta d},
\underset{t\rightarrow t-1}{\Delta x^L}, 
\underset{t\rightarrow t-1}{\Delta y}\}$, which is a backward flow defined on $D \times H \times W$ dimensions of the current cost volume. $\underset{t\rightarrow t-1}{\Delta d}\in \mathbb{R}^{D \times H \times W}$ and $\underset{t\rightarrow t-1}{\Delta x^L}, 
\underset{t\rightarrow t-1}{\Delta y}\in \mathbb{R}^{H \times W}$ because all $\delta$ in a spatial coordinate $(i,j)$ shares a common spatial flow.
The concept diagram of 3D flow $f_t^C$
can be found on the left side of Fig.~\ref{fig:cost_warp}. For understanding, arrows are aligned to the forward direction only within the figure. 

In ordinary cases, it is impossible to estimate the disparity flow $\Delta d$ for all disparity candidates $\delta \in [1,D]$ since there is no clue of where an arbitrary disparity candidate $\delta_{t-1}$ would be in time $t$. However, as a stereoscopic flow defines the independent flow for all coordinates in left and right event voxel in our model, we can define a relation between disparity flow and stereoscopic flow. For arbitrary point $(i_{t-1}, j_{t-1})$ on a left frame and a matching point with arbitrary disparity $\delta_{t-1}$ on a right frame, which is located in $(i_{t-1}-\delta_{t-1}, j_{t-1})$, we can track the movement of both points across time with stereoscopic flow. Knowing the locations of two points in $t-1$ and $t$, refers that we know the disparity changes for arbitrary coordinates and disparities across time. The right side of Fig.~\ref{fig:cost_warp} illustrates that the change of disparity can be expressed as the difference of $\Delta x^L$ and $\Delta x^R$, which can be written as Eq.(\ref{equ:disp_flow}).
For clarity, all flows are defined as backward flows to use backward warping.

\begin{equation}
\begin{aligned}
\underset{t\rightarrow t-1}{\Delta d}(\delta_t, i_t, j_t)
=\underset{t\rightarrow t-1}{\Delta x^L}(i_t, j_t)
-\underset{t\rightarrow t-1}{\Delta x^R}(i_t-\delta_t, j_t)
\label{equ:disp_flow}
\end{aligned}
\end{equation}
\noindent

At every depth estimation time step, the previous time cost volume $C_{t-1}$ is aligned to the current cost volume in three dimensions $D, H,$ and $W$, using the 3D backward flow $f^c_t$. 3-dimensional backward warping is applied to the previous cost volume, resulting in a temporally aligned cost $\tilde{C_t}$, as depicted in Eq.(\ref{equ:cost_warp}). This novel approach allows us to fully utilize past stereo knowledge. As we recycle the stereoscopic flow, the only additional computational cost is the warping function itself.
\begin{equation}
\begin{aligned}
\mathcal{W}_c&(C_{t-1},\,f_t^C)(c,\,d,\,h,\,w) \\
&=C_{t-1}(c,\ d+\underset{t\rightarrow t-1}{\Delta d}(d, \,h, \,w), \
h +\underset{t\rightarrow t-1}{\Delta y}(h,\,w), \ w+ \underset{t\rightarrow t-1}{\Delta x^L}(h,\,w))
\label{equ:cost_warp}
\end{aligned}
\end{equation}

Temporally aligned cost volume $\tilde{C_t}$ is selectively fused with the current cost volume $C_t$ based on their reliability. To balance computational cost and performance, we use the reliability filter module, which is shown to be useful in \cite{zhang2023vis, cheng2020deep,park2018learning, won2020end}. A disparity probability volume $P_t \in \mathbb{R}^{D \times H \times W}$ is calculated from the initial cost volume $C_t^{init}$ by channel aggregation block, which is composed of 3D convolutions. Then, the entropy map is calculated as the Eq.(\ref{equ:entropy}).
\begin{equation}
\begin{aligned}
E_t(i,j) = \sum_{d=0}^{D-1}-p_{i,j} \cdot \log{p_{i,j}}
\label{equ:entropy}
\end{aligned}
\end{equation}
\noindent The entropy map of the previous cost, $E_{t-1} \in \mathbb{R}^{H \times W}$, is calculated in the same way and warped to the current time frame with the warping function $\mathcal{W}_s$ and flow $f_t^L$. The previous and the current entropy map is concatenated and processed with a few 2D convolution layers to calculate the pixel-wise weight of two cost volumes. Weighted sum of $C_t$ and $\tilde{C}_t$ finalizes the cost fusion process.


\subsection{Temporal Disparity Consistency Loss} 

We propose the temporal disparity consistency (TDC) loss to provide a supervision signal for the stereoscopic flow network without extra flow ground truth. At a training phase, a previous GT disparity is warped to a current frame, compared with a current ground truth. 
When warping a disparity map, there should be a change in disparity value in a disparity map due to the relative pose of the camera and object as well as spatial coordinate shift.
Therefore, the change in disparity value, which we define as residual disparity, is to be added to the warped disparity. As shown in the right side of Fig.~\ref{fig:cost_warp}, the residual disparity between two time steps can be calculated as the difference in pixel movement along the horizontal line between matching points in stereo cameras, which can be derived from the stereoscopic flow. 
Specifically, we apply the backward function $\mathcal{W}_s$ to $\underset{t\rightarrow t-1}{\Delta x^R}$, which is defined in the right frame coordinate, to align with the left frame coordinate where disparity and cost volume are defined.
Then, the residual disparity, $\Delta D_{t,gt}$, between $t-1$ and $t$ can be calculated using current disparity ground-truth, $D_{t,gt}$, as:
\begin{equation}
\begin{aligned}
\Delta D_{t,gt}
=\mathcal{W}_s(\underset{t\rightarrow t-1}{\Delta x^R}, \{-D_{t,gt}, 0\})
-\underset{t\rightarrow t-1}{\Delta x^L}
\label{equ:residual_disp}
\end{aligned}
\end{equation}
The previous GT disparity $D_{t-1,gt}\in \mathbb{R}^{H \times W}$ can be aligned to the current time step using left camera flow components $\underset{t\rightarrow t-1}{\Delta x^L}, \underset{t\rightarrow t-1}{\Delta y}$ from  $\mathcal{SF}_t$. As shown in Eq.(\ref{equ:disparity_warp}), the first term is in charge of spatial pixel movement, while the residual disparity term deals with the disparity value change of each pixel.

\begin{equation}
\begin{aligned}
\Tilde{D}_{t,gt}
=\mathcal{W}_s(D_{t-1,gt}, \{\underset{t\rightarrow t-1}{\Delta x^L}, \underset{t\rightarrow t-1}{\Delta y}\}) 
+\Delta D_{t,gt}
\label{equ:disparity_warp}
\end{aligned}
\end{equation}
\noindent
We define the error between current GT disparity and warped previous GT disparity as the TDC loss using smooth L1 loss~\cite{girshick2015fast} as:
\begin{equation}
\begin{aligned}
\mathcal{L}_{tdc}= \mathrm{Smooth}_{L_1}(D_{t,gt},\tilde{D}_{t,gt})
\label{equ:loss_tdc}
\end{aligned}
\end{equation}

\subsection{Objective Functions}
\label{sec:loss_functions}
\noindent
\textbf{Flow Loss.}
We utilize the proposed \(\mathcal{L}_{tdc}\) (Eq.(\ref{equ:loss_tdc})) to optimize stereoscopic flow from events along with stereo matching loss, even in the absence of ground truth for optical flow. For better flow estimation on edges, contrast maximization loss~\cite{paredes2020back, zhu2019unsupervised}, $\mathcal{L}_c$, is also used as follows:
\begin{equation}
\begin{aligned}
\mathcal{L}_{flow}= \lambda_{t} \cdot \mathcal{L}_{tdc} + \lambda_{c}\cdot\mathcal{L}_{c}(e_t)
\label{equ:loss_flow}
\end{aligned}
\end{equation}

\noindent
\textbf{Stereo Matching Loss. } There are two pseudo predictions that are only inferred in training time using the current event, and a temporally refined final prediction. Smooth L1 loss is applied for three predictions.
\begin{equation}
    \begin{aligned}\mathcal{L}_{stereo}=\sum\limits_{i\in{0, 1}}\lambda_i \cdot \mathrm{Smooth}_{L_1}({D}_t^i, D_{t,gt}) +\lambda_f \cdot \mathrm{Smooth}_{L_1}(D_t, D_{t,gt})
    \end{aligned}
\end{equation}

In summary, the total objective function $\mathcal{L}$
is a summation of flow loss ($\mathcal{L}_{flow}$) and stereo loss ($\mathcal{L}_{stereo}$) for joint learning with stereoscopic flow as:
\begin{equation}
\begin{aligned}
\mathcal{L}= \mathcal{L}_{flow} + \mathcal{L}_{stereo}
\label{equ:loss_total}
\end{aligned}
\end{equation}

\section{Experiments}

\subsection{Datasets}
Our model is verified in MVSEC dataset~\cite{8288670} following the previous works. The dataset consists of event streams from calibrated stereo event cameras, which have a resolution of 346$\times$260, and ground truth depth. 
For a fair comparison, we adhered to the protocol~\cite{tulyakov2019learning, zhang2022discrete, ahmed2021deep} by conducting training and evaluation on sequences 1 and 3 of the Indoor Flying scene.
For the metric, mean depth error, median depth error, mean disparity error, and one-pixel-accuracy (1PA) are used. Computational complexity is measured in terms of frame rate and FLOPs. 

Furthermore, to validate the generalizability of the model, experiments are conducted on DSEC dataset~\cite{gehrig2021dsec}. It consists of 53 different driving scenes with stereo event cameras of 640$\times$480 resolution. One-pixel-error (1PE), two-pixel-error, mean absolute error (MAE), and root mean square error (RMSE) are measured following the previous works and the benchmark website.

\begin{table*}[t]
\centering
\caption{Results for event-based stereo on MVSEC and DSEC datasets. * indicates that images were also used in the training. - indicates that official results are not provided in the original paper. The FPS indicates the GPU synchronized inference time on the MVSEC, using an NVIDIA TITAN Xp. The FPS measures are an average of 1000 iterations after 100 warm-up loops. ``Sp'' is an abbreviation for ``split''.}
\resizebox{1.0\textwidth}{!}{
\setlength\tabcolsep{4.3pt}
\begin{tabular}{l||c|c|c|c|c|c|c|c||c|c|c|c||c|c}
\thickhline 
\multirow{5.1}{*}{ Method }  & \multicolumn{8}{c||}{MVSEC} & \multicolumn{4}{c||}{DSEC} & \multicolumn{2}{c}{Cost}\\
\cline { 2 - 15 }
& \multicolumn{2}{c|}{\thead{Mean depth \\ error [cm] $\downarrow$}} & \multicolumn{2}{c|}{\thead{Median depth \\ error [cm] $\downarrow$}} & \multicolumn{2}{c|}{\thead{Mean disp. \\ error [pix] $\downarrow$}} & \multicolumn{2}{c||}{ 1PA [\%] $\uparrow$} & \multirow{0.5}{*}{\thead{ 1PE \\ $\downarrow$}} & \multirow{0.5}{*}{\thead{ 2PE \\ $\downarrow$}} & \multirow{0.5}{*}{\thead{ MAE \\ $\downarrow$}} & \multirow{0.5}{*}{\thead{ RMSE \\ $\downarrow$}} & \multirow{2.0}{*}{\thead{ FPS \\ $\uparrow$}} & \multirow{2.0}{*}{\thead{ GFlops \\ $\downarrow$}} \\
\cline { 2 - 9 } & Sp 1 & Sp 3 & Sp 1 & Sp 3 & Sp 1 & Sp 3 & Sp 1 & Sp 3  & & & & & &  \\ 
\thickhline 
DDES~\cite{tulyakov2019learning} & 16.7 & 27.8 & 6.8 & 14.7 & 0.59 & 0.94 & 89.8 & 74.8 & 10.915 & 2.905 & 0.576 & 1.381 & $\underline{34.3}$ & 75.8  \\
DTC-PDS~\cite{zhang2022discrete} & 15.3 & 18.6 & 6.4 & 8.7 & 0.56 & 0.65 & 91.5 & 88.7 & 9.517 & \underline{2.356} & 0.527 & 1.264 & 33.2 & \underline{63.9}\\
CTC-PDS~\cite{zhang2022discrete} & 14.9 & 20.6 & 6.4 & 10.6 & \underline{0.53} & 0.73 & 91.6 & 88.2 & - & - & - & - & - & - \\
EITNet*~\cite{ahmed2021deep} & 14.2 & 19.4 & 5.9 & 10.4 & 0.55 & 0.75 & 92.1 & 89.6 & - & -& - & -& - & -\\
Se-CFF~\cite{nam2022stereo} & - & - & - & - & - & - & - & - & 9.583 & 2.620 & \underline{0.519} & 1.231 & 22.2 & 139.4 \\
EIS-E~\cite{mostafavi2021event} & \underline{13.3} & 25.7 & - & - & - & - & 80.6 & 68.3 & 9.958 & 2.645 & 0.529 & \underline{1.222} &- & - \\
DTC-SPADE~\cite{zhang2022discrete} & 13.5 & \underline{17.1} & $\underline{5.2}$ & \underline{7.9} & $\mathbf{0.46}$ & \underline{0.60} & $\mathbf{93.0}$ & \underline{89.7} & \underline{9.277} & 2.416 & 0.527 & 1.291 & 11.6 & 237.4\\
\hline
\textbf{Ours} & $\mathbf{13.0}$ & $\mathbf{15.0}$ & $\mathbf{5.0}$ & $\mathbf{5.8}$ & $\mathbf{0.46}$ & $\mathbf{0.49}$ & \underline{92.9} & $\mathbf{92.6}$ & $\mathbf{8.662}$ & $\mathbf{2.259}$ & $\mathbf{0.493}$ & $\mathbf{1.172}$ & $\mathbf{71.6}$ & \textbf{57.4}\\ 
\thickhline 
\end{tabular}
}
\label{tab:main}
\end{table*}

\subsection{Implementation Details}

The models used in MVSEC and DSEC have identical structures, but we set the maximum disparity to 48 and 192, respectively. Additionally, to accommodate the disparity range, we increased the feature encoder channels. More details on implementation are given in the supplementary material.

\begin{figure*}[t]
\begin{center}
\includegraphics[width=.88\linewidth]{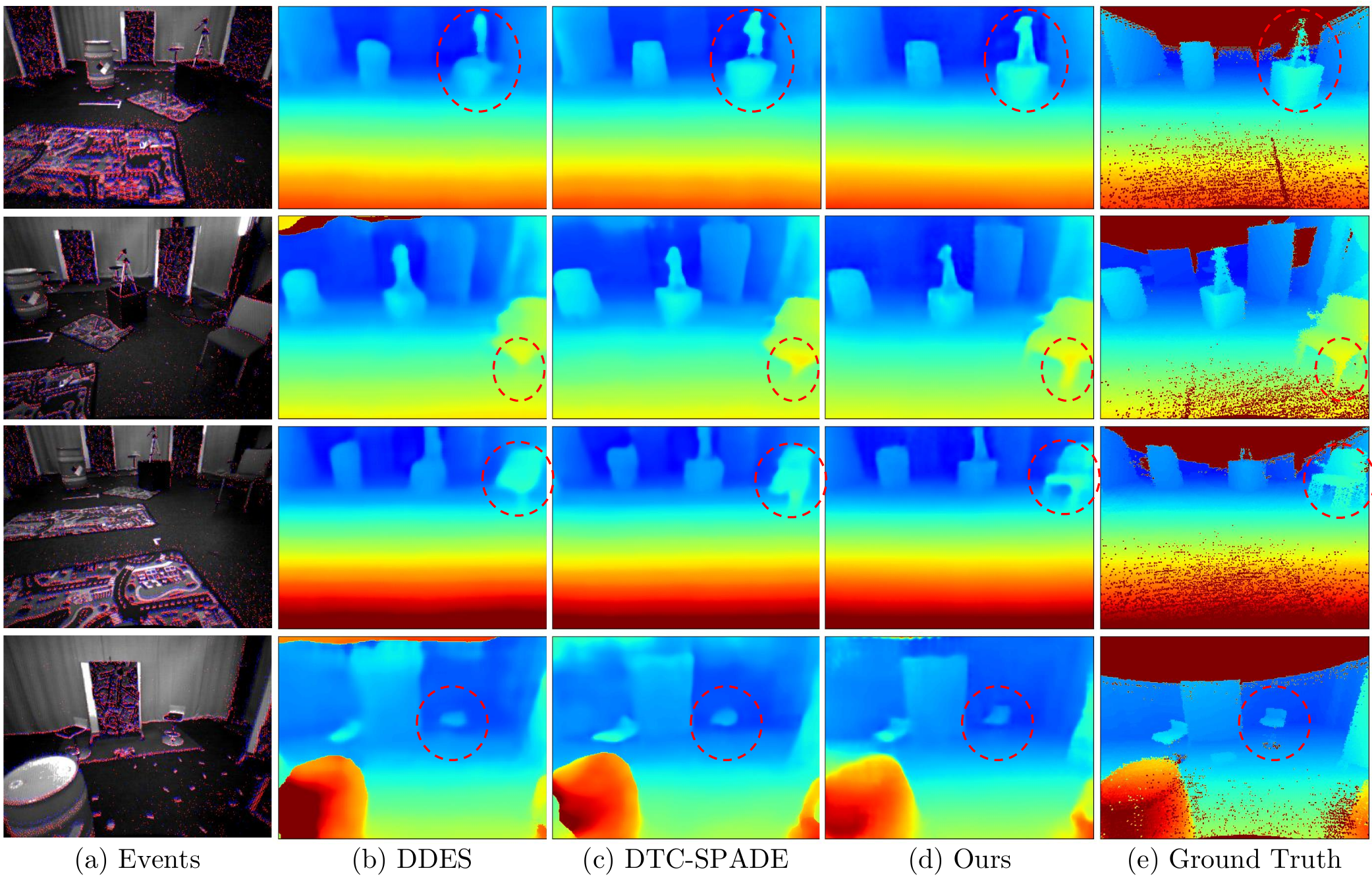}
\caption{Qualitative results on the Indoor Flying dataset of MVSEC. From top to bottom, the rows display $\#245$ from sequence 1, $\#591$ from sequence 1, $\#200$ from sequence 3, and $\#1585$ from sequence 3, respectively. The author reproduced qualitative results with publicly available code~\cite{zhang2022discrete, tulyakov2019learning}.}
\label{fig:qual_mvsec}
\end{center}
\end{figure*}

\begin{figure*}[t]
\begin{center}
\includegraphics[width=.88\linewidth]{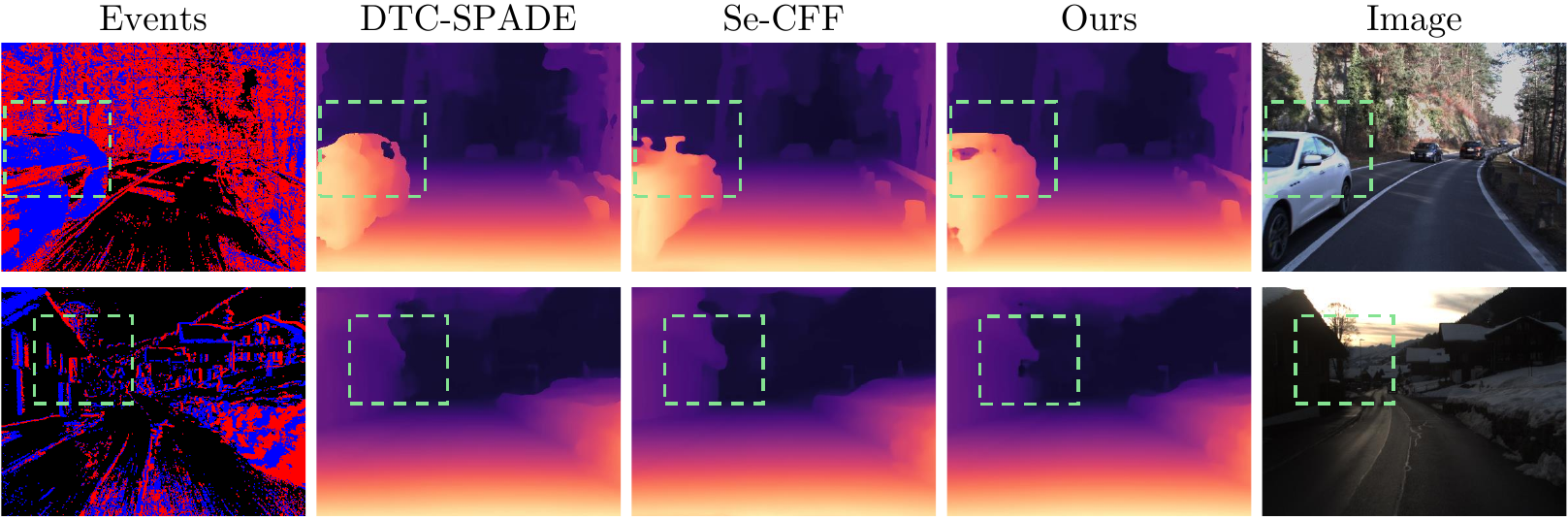}
\caption{Qualitative results on the DSEC dataset. We use a pre-trained model provided by the author for Se-CFF~\cite{nam2022stereo} and train DTC-SPADE~\cite{zhang2022discrete} using public code.}
\label{fig:qual_dsec}
\end{center}
\end{figure*}

\subsection{Quantitative Results}
Table~\ref{tab:main} shows the dense disparity estimation results on both MVSEC and DSEC datasets. Focusing on precise depth estimation performance only using asynchronous and sparse events, the table compares event-only stereo methods tested either on MVSEC or DSEC. 
The results of other methods on MVSEC are taken from their original paper, and the results of DSEC are derived from the DSEC benchmark. 
The DTC-SPADE~\cite{zhang2022discrete} achieved remarkable performance outperforming EITNet~\cite{ahmed2021deep} which uses images in training, by introducing RNN structure for feature embedding. However, as a price of temporal encoding, computational cost increased significantly. 
Our main model outperforms previous SOTA except 1PA, while minimizing computation cost. A lightweight stereoscopic flow network and warping process enable our model to propagate stereoscopic knowledge across time with minimal burden. Since enough information is passed from the previous time step, a small feature encoder is sufficient to make a precise prediction.
For the results on DSEC dataset, our results are uploaded and evaluated on the DSEC benchmark website\footnote{\url{https://dsec.ifi.uzh.ch/uzh/disparity-benchmark/}}. Our method achieves the best performance across all metrics and outperforms other methods by a significant margin.

\subsection{Qualitative Results}
Qualitative results of the proposed model and the previous event stereos are shown in Fig.~\ref{fig:qual_mvsec} and Fig.~\ref{fig:qual_dsec}. DDES~\cite{tulyakov2019learning} and Se-CFF~\cite{nam2022stereo}, single pair event-based stereo matchings, 
often fail to reconstruct scenes due to a lack of sufficient event information.
In contrast, DTC-SPADE~~\cite{zhang2022discrete}  shows better qualitative results and stable depth reconstruction across the entire scene thanks to temporally aggregated features, yet it still struggles with edges and boundaries.
Conversely, our method can compensate for the lack of feature information and 3D cues from past cost volumes, demonstrating remarkably sharp qualitative results for objects.

\subsection{Ablation Study and Analysis}
\noindent
\textbf{Ablation Study on Temporal Modules.}
A performance gain of each module and the effect of temporal aggregation are shown in Table~\ref{tab:ablation_modules}. The baseline is a single-frame implementation of our model, which does not use past knowledge. Adding a feature warping module improves performance to a sub-optimal level by enriching the representation of each scene. Also, adding the cost volume warping module improved all metrics by a large margin, by providing dense and abstract knowledge related to stereo matching to the next time step.
Finally, the best performance is achieved when both scene-specific feature information and stereo-centric costs are utilized.

\noindent
\textbf{Reliability Filter.}
There can be many different methods to fuse two separate cost volumes from different time sources. As a cost volume represents a matching score for each disparity candidate, naive approaches like addition or channel-wise concatenation do not exploit the benefit of cost volume warping. 
Therefore, to achieve optimal performance, we applied cost volume fusion using various approaches, as shown in Table~\ref{tab:ablation_filter}.
The addition method averages two cost volumes, while concatenation uses a convolution layer to reduce the channel. The cost filter method employs a pixel-wise weight estimation CNN block that takes concatenated cost volumes as input.
Among various fusion methods, the entropy filter achieves the best performance, so we adopt this approach.

\begin{table}[t]
    \begin{minipage}[h]{.62\linewidth}
        \centering
        \captionof{table}{Ablation studies of the proposed modules.}
        \renewcommand{\arraystretch}{1.1}
        \setlength\tabcolsep{4.2pt}
        \resizebox{0.99\linewidth}{!}{
        \begin{tabular}{l|c|c|c}
        \thickhline 
        Ablation Settings & \thead{Mean depth\\error [cm] $\downarrow$} & \thead{Mean disp\\error [pix] $\downarrow$} & 1PA [\%] $\uparrow$  \\
        \thickhline 
        Single & 15.2 & 0.53 & 91.4 \\ 
        + Feature Warping & 14.7 & 0.50 & 91.9 \\
        + Cost Volume Warping & \underline{13.6} & \underline{0.49} & \underline{92.3} \\ 
        \textbf{Full Model} & \textbf{13.0} & \textbf{0.46} & \textbf{92.9}  \\
        \thickhline
        \end{tabular}
        }
        \label{tab:ablation_modules}
    \end{minipage}
    \begin{minipage}[h]{.36\linewidth}
        \centering
        \captionof{table}{The ablation study of reliability filter.}
        \renewcommand{\arraystretch}{1.1}
        \resizebox{0.99\linewidth}{!}{
            \begin{tabular}{l|c|c|c}
        \thickhline 
        \multirow{2}{*}{}{\thead{Settings}} &
              \thead{Mean \\ Depth $\downarrow$} &
              \thead{Mean \\ Disparity $\downarrow$} &
              \thead{1PA $\uparrow$}  \\
        \thickhline 
        Add & 14.0 & 0.48 & 92.6 \\
        Concat & \underline{13.5} & \underline{0.47} & \underline{92.7}  \\
        Cost Filter & \underline{13.5} & \underline{0.47} & 92.6 \\
        \textbf{Ours} & \textbf{13.0} & \textbf{0.46} & \textbf{92.9} \\
        \thickhline
        \end{tabular}
        }
        \label{tab:ablation_filter}
        
    \end{minipage}
\end{table}

\noindent
\textbf{Stereoscopic Flow.}
We conducted ablations in two ways, regarding the effect and efficiency of the proposed stereoscopic flow network. First, to validate the effect of stereo feature sharing, we fed event voxels directly to a large flow network, EV-Flow~\cite{zihao2018unsupervised}, capable of high-quality flow estimation. Second, we made a twin optical flow network that separately estimates left and right optical flows. Smooth L1 loss is applied to constrain vertical flows, and it is denoted as soft epipolar constraint in Table~\ref{tab:ablation_sfnet}. Even if the added soft constraints trained two independent flow networks to estimate the same vertical flow, adding a redundant degree of freedom makes training inefficient. Our design, which uses a single stereoscopic flow network sharing stereo feature, outperforms other settings.

\noindent
\textbf{Temporal Disparity Consistency Loss.}
Table~\ref{tab:flow_loss} shows the ablation study of the proposed temporal disparity consistency loss. As explored in previous studies~\cite{gallego2018unifying}, event-based optical flow can be learned with just contrast maximization \(\mathcal{L}_c\), but joint training with other tasks can be unstable~\cite{shiba2022secrets} and potentially conflict with those tasks. In contrast, our \(\mathcal{L}_{tdc}\) is not a fully self-supervised learning approach and is optimized by cues from disparity, allowing for stable training. \(\mathcal{L}_{tdc}\) alone can achieve satisfactory performance, but the best performance is attained when trained together with contrast.

\begin{table}[t]
    \begin{minipage}[h]{.58\linewidth}
        \centering
        \captionof{table}{The analysis of the stereoscopic flow.}
        \setlength\tabcolsep{3.5pt}
        \renewcommand{\arraystretch}{1.15}
        \resizebox{0.99\linewidth}{!}{
            \begin{tabular}{l|c||c|c|c|c}
            \thickhline 
             \multirow{2}{*}{Settings} & Epipolar &
                  Mean  &
                  Mean  &
                   \multirow{2}{*}{1PA  $\uparrow$} &
            \multirow{2}{*}{FPS $\uparrow$} \\
                & Constraint & Depth $\downarrow$ & Disp $\downarrow$ & \\
            \thickhline 
            EV-Flow~\cite{zhu2018ev} & Soft &14.1 & 0.51 & 91.8 &  38.7\\
            EV-Flow~\cite{zhu2018ev} & Hard $(\mathcal{SF})$ & 13.7 & 0.49 & \underline{92.2} &  51.0\\ \hline
            Ours & Soft & \underline{13.3} & \underline{0.47} & \textbf{92.9} &  \underline{69.7}\\
            \textbf{Ours} & Hard $(\mathcal{SF})$ & \textbf{13.0} & \textbf{0.46} & \textbf{92.9}  & \textbf{71.6}\\
            \thickhline
            \end{tabular}
            }
        \label{tab:ablation_sfnet}
    \end{minipage}
    \hfill
    \begin{minipage}[h]{.4\linewidth}
        \centering
        \captionof{table}{The ablation study of flow loss (Eq.~\ref{equ:loss_flow}).}
        \resizebox{0.99\linewidth}{!}{
        \setlength\tabcolsep{4.7pt}
        \begin{tabular}{l|c|c|c}
        \thickhline 
        \multirow{2}{*}{}{\thead{Ablation \\ Settings}} &{\thead{Mean \\Depth $\downarrow$}} &
          \thead{Mean \\ Disp $\downarrow$} &
          \thead{1PA $\uparrow$} \\
        \thickhline 
        $\mathcal{L}_{c}$ & 13.6 & \underline{0.47} & \underline{92.4} \\
        $\mathcal{L}_{tdc}$ & \underline{13.1} & \underline{0.47} & \textbf{92.9} \\
        $\mathcal{L}_{c}+\mathcal{L}_{tdc}$ & \textbf{13.0} & \textbf{0.46} & \textbf{92.9} \\
        \thickhline
        \end{tabular}
        }
        \label{tab:flow_loss}
    \end{minipage}
\end{table}

\subsection{Performance Gain from Temporal Consistency}
\begin{figure*}[t]
\begin{center}
\includegraphics[width=.98\linewidth]{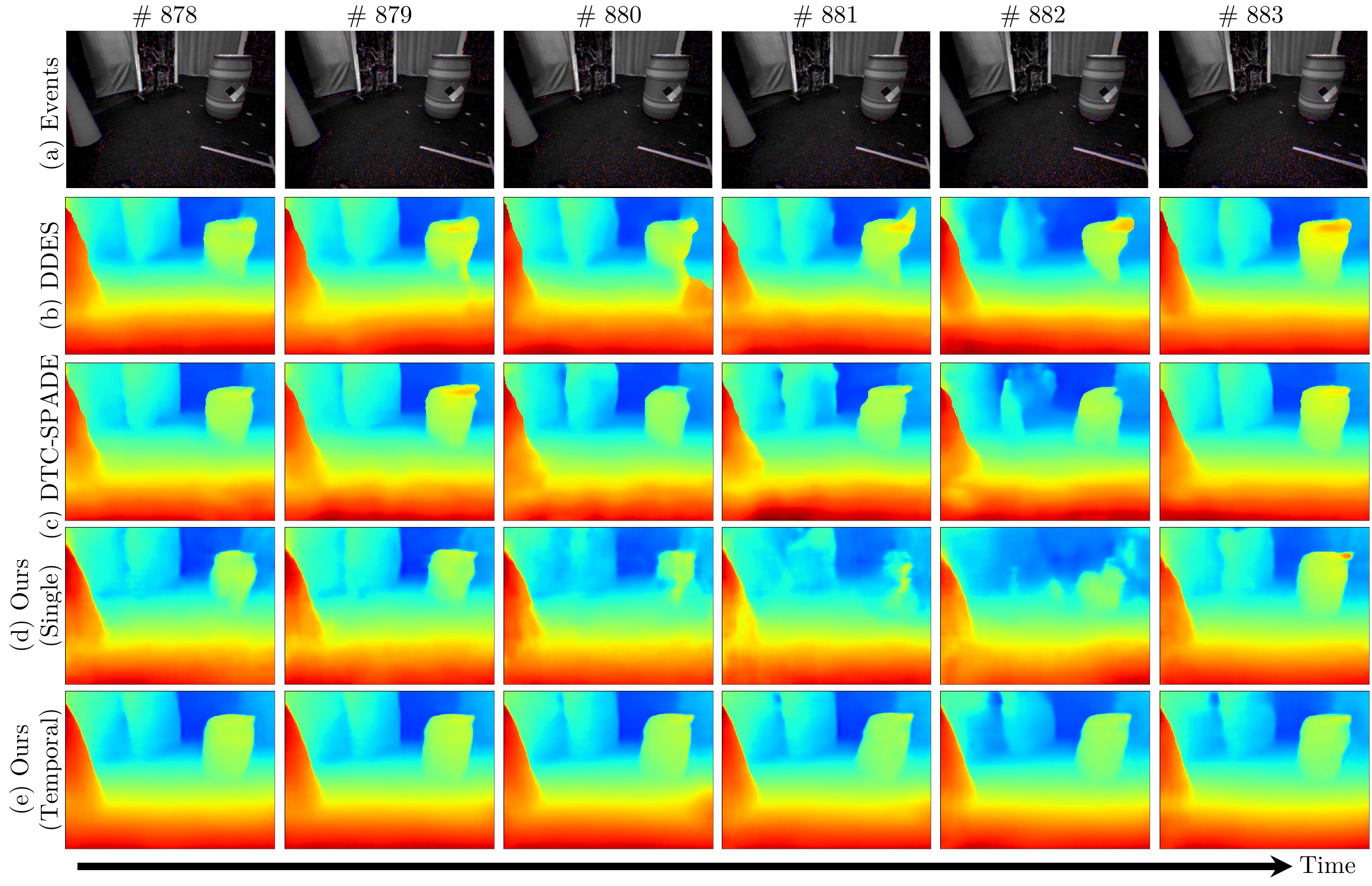}
\caption{Qualitative results for temporal consistency on split 1 of MVSEC dataset.}
\label{fig:qual_consistency}
\end{center}
\end{figure*}

To further validate the effectiveness of temporal aggregation, we provide Fig.~\ref{fig:qual_consistency}, which shows the sequential qualitative results to deliver intuition of the impact of knowing the past. Due to the small motion, event data lacks information on the overall scene, \eg,~cylinder object. The single frame estimation methods, DDES and our single baseline, fail to estimate disparity in some time steps that lack event streams. DTC-SPADE estimates rough disparity for challenging scenes but lacks edge details due to the limited temporal aggregation. Our model estimates consistent disparities using past stereo information, which is the behavior to 
improve stereo performance reflecting the real world.

\section{Conclusion}
We propose a new paradigm of a temporal event stereo network that aggregates past features and cost volume to enhance stereo representations using a stereoscopic flow, which is a stereo-targeted novel flow representation. Previous features and cost volume are aligned to the current time step, enriching the stereo representations. Especially, our model warps the entire cost volume, a stereo-specific representation, across the time frame achieving significant performance improvement with minimal computational cost. We also propose a TDC loss and use encoder-sharing architecture to train stereoscopic flow and stereo jointly without using flow GT. From this effort, we achieve state-of-the-art performance in both MVSEC and DSEC datasets with minimal computation.

\noindent
\textbf{Limitations and Future Work.} 
Our framework demonstrates training stereo flow without ground truth, but self-supervised stereo matching using events alone remains unexplored. Our approach could inspire future research in this area.

\noindent
\textbf{Acknowledgements.} This work was supported by the Technology Innovation Program (1415187329,20024355, Development of autonomous driving connectivity technology based on sensor-infrastructure cooperation) funded By the Ministry of Trade, Industry \& Energy(MOTIE, Korea) and the National Research Foundation of Korea(NRF) grant funded by the Korea government(MSIT) (NRF2022R1A2B5B03002636).

%
%
\bibliographystyle{splncs04}
\bibliography{reference_bib}

 



\title{Temporal Event Stereo via Joint Learning with Stereoscopic Flow\\
\textmd{Supplementary Material}} 

\author{Hoonhee Cho\orcidlink{0000-0003-0896-6793}\thanks{Equal contribution.} \and
Jae-Young Kang\orcidlink{0009-0002-9537-3813}\protect\CoAuthorMark \and
Kuk-Jin Yoon\orcidlink{0000-0002-1634-2756}}

\authorrunning{Cho et al.}

\institute{Korea Advanced Institute of Science and Technology\\
\email{\{gnsgnsgml, mickeykang, kjyoon\}@kaist.ac.kr}\\
}

\maketitle
In this supplementary material, we provide more details that are not included in the paper due to space limitations.
This includes the conceptual comparisons (Sec.~\ref{sec:concept}), implementation details (Sec.~\ref{sec:details}), and more experimental results and analyses (Sec.~\ref{sec:results_analyses}), respectively.

\setcounter{figure}{0}
\setcounter{table}{0}
\setcounter{algorithm}{0}
\setcounter{equation}{0}
\renewcommand{\thefigure}{A\arabic{figure}}
\renewcommand{\thetable}{A\arabic{table}}
\renewcommand{\thealgorithm}{A\arabic{algorithm}}
\renewcommand{\theequation}{A\arabic{equation}}
\renewcommand{\thesection}{A}

\section{Conceptual Comparisons}
\label{sec:concept}
We provide a conceptual comparison to highlight the differences from existing methods. Se-CFF~\cite{nam2022stereo} fuses multi-density event stacks within a single timeframe.
Other methods~\cite{zhang2022discrete, mostafavi2021event} use recurrent architecture to aggregate feature-level temporal information. 
As shown in Table~\ref{tab:method_rebuttal}, conceptually,  our method explicitly retrieves temporal information based on flow rather than relying on an implicit feature-level approach. This approach offers the advantage of reusing the computed flow in both feature and cost volume. Furthermore, a key difference lies in our cost volume aggregation, which is essential and task-dependent for stereo matching.

\begin{table}[h]
    \centering
    \caption{The summary of the conceptual differences.}
    \setlength\tabcolsep{8.5pt}
    \renewcommand{\arraystretch}{1.15}
    \resizebox{0.80\linewidth}{!}{
        \begin{tabular}{l|c|c|c}
        \thickhline 
         \multirow{3}{*}{Method} & \multicolumn{2}{c|}{Target Temporal}
         & Temporal 
         \\ 
         & \multicolumn{2}{c|}{Information} & Retreieval \\
         \cline{2-3} &  Feature & Cost Volume &  Method 
         \\ \cline{1-4} 
         DDES~\cite{tulyakov2019learning}, EITNet~\cite{ahmed2021deep} &  $\tikzxmark$ & $\tikzxmark$ &  $\tikzxmark$
         \\  
         Se-CFF~\cite{nam2022stereo} &  $\triangle$ & $\tikzxmark$ & CNN-based
         \\ EIS-E~\cite{mostafavi2021event}, 
         DTC~\cite{zhang2022discrete} & $\tikzcmark$ & $\tikzxmark$ & RNN-based
         \\ \textbf{Ours} & $\tikzcmark$ & $\tikzcmark$ & Flow-based\\
         \thickhline
        \end{tabular}
        }
    \label{tab:method_rebuttal}
\end{table}

\setcounter{figure}{0}
\setcounter{table}{0}
\setcounter{algorithm}{0}
\setcounter{equation}{0}
\renewcommand{\thefigure}{B\arabic{figure}}
\renewcommand{\thetable}{B\arabic{table}}
\renewcommand{\thealgorithm}{B\arabic{algorithm}}
\renewcommand{\theequation}{B\arabic{equation}}
\renewcommand{\thesection}{B}

\section{Implementation Details}
\label{sec:details}

\subsection{Measurement of Inference Time}
As mentioned in the official implementations of \cite{cho2021rethinking, scaffoldgs}, measuring the inference time of PyTorch models requires avoiding the use of `time.time()' due to the asynchronous nature of GPU operations. Instead, two main steps must be followed to accurately measure the inference time:

\noindent
\textbf{1. GPU Warm-up.} Before measuring inference time, we should warm up the GPU by running some dummy examples through the network. This step is crucial because a GPU can exist in various power states, and without warm-up, the device might not operate at its full capacity during the initial runs, leading to inaccurate timing measurements.

\noindent
\textbf{2. GPU and CPU Synchronization.} To ensure accurate timing measurements, it's essential to use `torch.cuda.synchronize()' before and after the inference calls. This function synchronizes the CPU and GPU, ensuring that all GPU tasks are completed before the time is measured. This step overcomes the potential inaccuracies caused by the asynchronous execution of GPU operations.

For accurate and fair measurement of FPS, we followed the mentioned methods. Additionally, we performed a GPU warm-up using 100 dummy examples to ensure the GPU was fully operational for the actual measurements.

\subsection{Network Architecture}

The network structure and parameters for the MVSEC dataset are detailed in Table~\ref{tab:Structure_Tables}. We designed the network to be efficient by significantly reducing the channel dimension of features while allowing for temporal aggregation. For DSEC, we increased the channel dimensions of the multi-scale in Table~\ref{tab:Structure_Tables} from 12, 24, and 36 to 32, 64, and 128, respectively.

\subsection{More Implementation Details}

In the MVSEC dataset, event streams are sliced every 50ms and processed to voxels~\cite{zhu2019unsupervised} of bin 5. For flow loss weights, $\lambda_t=0.1$ and $\lambda_c=1 \cdot 10^{-5}$ are used. For stereo loss, $\lambda_0=0.5$, $\lambda_1=0.7$, and $\lambda_f=1.0$. The maximum disparity of cost volume is set to 48. 
In the train phase, four serial event voxel pairs are sequentially fed into the model, and only the last stereo pair is used for loss calculation, while previous frames are only inferenced for temporal information. Training is done in an end-to-end manner for 60 epochs and a batch size of 2. The learning rate is set to 0.0008 using Adam optimizer~\cite{kingma2014adam}. 

Faster ego-motion in DSEC tends to trigger more events per unit of time. Therefore, each 50ms event stream is processed to bin 15 voxel. Also, flow loss weights are reduced to $\lambda_t=0.01$ and $\lambda_c=1 \cdot 10^{-8}$ for stable training. The maximum disparity is set to 192. The model is trained for 100 epochs with batch size 4.

During training, we grouped 4 sequential event voxel pairs into one clip for a training purpose, and similarly, for testing, we also used clips consisting of 4 voxels each. Additionally, to make the temporal disparity consistency more stable, we adopted the approach of pseudo ground-truth (GT), a method previously utilized in stereo research~\cite{xu2020aanet}. For this purpose, we first trained a single event stereo network without any temporal aggregation and performed inference of disparity maps for pseudo-GT. In areas where sparse GT was available, we used the GT, and in areas without GT, we filled in with pseudo-GT. The densified GT disparity map is only used for flow loss calculation, not for the stereo loss.

In the main paper, our explanation focuses on cost volume warping and temporal disparity consistency loss. 
Due to the epipolar constraint, there is no need to consider \(\Delta y^R\) in cost volume warping and temporal disparity consistency, so we omitted  \(\Delta y^R\) for a better explanation of our core ideas. In actual implementation, \(\Delta y^R\) is also estimated with other components by the single stereoscopic flow network and only used for feature warping, as mentioned in the main paper.

\subsection{Details About the Ablation Study in Table 4 of the Main Paper}
In the main paper, we present Table 4 for the ablation study of stereoscopic flow. Table 4 aims to validate stereoscopic flow from two perspectives.

First, it validates the impact of sharing stereo features by comparing outcomes between sharing and not sharing stereo features as inputs to the flow network. Instead of sharing stereo features, we employ the comprehensive optical flow network, EV-Flow~\cite{zhu2018ev}, with event voxel input. In the table, ``ours'' signifies feature sharing with a stereo network, and ``EV-Flow'' denotes a flow network independent of stereo features.

Second, we validate the relationship between the left and the right flows in stereo matching. In other words, we tested the assumption of the `hard' epipolar constraint that matching points share the same vertical flow. In a hard constraint setup, which is our baseline, both left and right event information are fed into a single stereoscopic flow network, estimating 4-dimensional flow for stereo. Also, cost volume warping and temporal disparity consistency (TDC) loss are calculated based on the assumption that matching pixels share the same vertical flow. In contrast, in soft epipolar constraint experiments, the left and the right flows are estimated independently and calculated separately. Two twin flow networks are used for each left and right flow. $\{
\underset{t\rightarrow t-1}{\Delta x^L},
\underset{t\rightarrow t-1}{\Delta y^L}\}$ are estimated from using only left event while $\{
\underset{t\rightarrow t-1}{\Delta x^R},
\underset{t\rightarrow t-1}{\Delta y^R}\}$ are predicted only from right events. Moreover, cost volume and TDC loss calculation are modified to accommodate different vertical flows. As a result, computational complexity is increased for an extra flow network. However, the redundant degree of freedom on vertical flows negatively affected the stereo matching performance.

\setcounter{figure}{0}
\setcounter{table}{0}
\setcounter{algorithm}{0}
\setcounter{equation}{0}
\renewcommand{\thefigure}{C\arabic{figure}}
\renewcommand{\thetable}{C\arabic{table}}
\renewcommand{\thealgorithm}{C\arabic{algorithm}}
\renewcommand{\theequation}{C\arabic{equation}}
\renewcommand{\thesection}{C}

\begin{table}[b]
\caption{The result according to hyper-paramter in Eq.~(8). $\lambda_t$ and $\lambda_c$ refer to the weight of temporal disparity consistency (TDC) loss and contrast loss, respectively.}
\label{tab:hyper}
\begin{center}
\renewcommand{\arraystretch}{1.2}
\resizebox{0.96\linewidth}{!}{\renewcommand{\tabcolsep}{4.5pt}
\begin{tabular}{c|cc|cc|cc|cc}
\hline
\multirow{2}{*}{$\lambda_t$ \textbackslash{} $\lambda_c$} & \multicolumn{2}{c|}{$10^{-6}$}
& \multicolumn{2}{c|}{$10^{-5}$ } 
 & \multicolumn{2}{c|}{$10^{-4}$}   
& \multicolumn{2}{c}{$10^{-3}$}        \\
\cline{2-9}
& Mean Depth & 1PA & Mean Depth & 1PA & Mean Depth & 1PA & Mean Depth & 1PA  \\ 
\hline
0.01 & \gradient{13.8} & \gradienttwo{92.6} & 
\gradient{13.4} & \gradienttwo{92.3} & 
\gradient{13.5} & \gradienttwo{92.4} &
\gradient{15.5} & \gradienttwo{90.1}
\\
0.1 & \gradient{13.1} & \gradienttwo{92.7}  &
\gradient{13.0} & \gradienttwo{92.9}  &
\gradient{14.4} & \gradienttwo{91.9}  &
\gradient{14.6} & \gradienttwo{91.3}
\\
1.0  & \gradient{13.3} & \gradienttwo{92.8}  & 
\gradient{13.5} & \gradienttwo{92.5}  & 
\gradient{14.3} & \gradienttwo{91.7}  &
\gradient{15.7} & \gradienttwo{90.5} 
\\ 
\hline
\end{tabular}}
\end{center}
\end{table}

\section{Additional Experimental Results and Analyses}
\label{sec:results_analyses}
\subsection{Hyper-parameter Analysis of Flow Loss (Eq. (8))}
The flow loss (Eq. (8)) consists of the temporal disparity consistency loss, proposed for jointly learning stereoscopic flow with stereo, and a minor contrast loss~\cite{zhu2019unsupervised} serving as an auxiliary loss.
Table~\ref{tab:hyper} provides results based on the coefficient of the loss in the flow loss.
Contrast loss becomes unstable when jointly trained with the stereo network, and setting the \(\lambda_c\) beyond a certain value leads to a significant decrease in the performance of the stereo network. On the other hand, our TDC loss is robust even when trained together with the stereo network, maintaining results within a certain range regardless of the scale of the weight becoming larger or smaller. 

\subsection{Streaming Experiment}
The main MVSEC experiments are only conducted with limited past information; 4 stereo frames at the test phase. 
However, in the real world, streams of events are continuously fed into the model. Therefore, we conducted additional experiments to validate the real-world application, and to verify the long-term information propagation. We inferred our model, which is trained with 4 sequential frames, with different numbers of stereo frames: 2, 4, 8, and streaming. In the streaming experiment, all event voxels are sequentially processed and the random test sets are evaluated. The results are provided in Table~\ref{tab:streaming}. Even if the model is trained only for 4 frames setting, it can retrieve information from further past frames to enhance current disparity prediction.

\begin{table}[t]
\label{tab:streaming}
\begin{center}
\renewcommand{\arraystretch}{1.2}
\caption{Streaming experiment with different train/test frame}
\resizebox{0.99\linewidth}{!}{\renewcommand{\tabcolsep}{4.5pt}
\begin{tabular}{c|cc|cc|cc|cc}
\hline
\multirow{2}{*}{train \textbackslash{} test} & \multicolumn{2}{c|}{2 frames}
& \multicolumn{2}{c|}{4 frames (Base)} 
 & \multicolumn{2}{c|}{8 frames}   
& \multicolumn{2}{c}{Streaming}        \\
\cline{2-9}
& Mean Depth & 1PA & Mean Depth & 1PA & Mean Depth & 1PA & Mean Depth & 1PA  \\ 
\hline
4 frames & \gradientfour{17.2}(4.2 $\uparrow$) & \gradientthree{89.9}(3.0 $\downarrow$)  &
\gradientfour{13.0} & \gradientthree{92.9} &
\gradientfour{12.8}(0.2 $\downarrow$) & \gradientthree{93.0} (0.1 $\uparrow$)&
\gradientfour{12.8}(0.2 $\downarrow$) & \gradientthree{93.0} (0.1 $\uparrow$)
\\
\hline
\end{tabular}}

\label{tab:streaming}
\end{center}
\end{table}

\subsection{Stereoscopic Flow}




Stereoscopic flow is an auxiliary output to facilitate temporal information from the past. Even if the quality and quantity of flow results are not our main interest, visualization of the intermediate outputs is useful for understanding the model behavior. The stereoscopic flow network generates 4-dimensional flow, 
$\{
\underset{t\rightarrow t-1}{\Delta x^L},
\underset{t\rightarrow t-1}{\Delta x^R},
\underset{t\rightarrow t-1}{\Delta y^L},
\underset{t\rightarrow t-1}{\Delta y^R}\}$, and we visualize the left camera flow $\{\underset{t\rightarrow t-1}{\Delta x^L}, \underset{t\rightarrow t-1}{\Delta y^L}\}$ among them in the Fig.~\ref{fig:qual_flow_mvsec}.
As the network estimates flow in 1/4 resolution of the input voxel grid, we applied bilinear upsampling for visualization.


\begin{table}[t]
\centering
\caption{Experimental results according to the bin size of the voxel grid.}
\resizebox{0.72\linewidth}{!}{
\setlength\tabcolsep{17.0pt}
\renewcommand{\arraystretch}{1.1}
\begin{tabular}{c|c|c|c}
\hline
Bin Size & Mean Depth $\downarrow$ &
      Mean disp $\downarrow$ &
      1PA $\uparrow$ \\
\hline
1 & 13.7 & 0.47 & 92.2\\
5 & \textbf{13.0} & \textbf{0.46} & \textbf{92.9}\\
10 & 13.7 & 0.48 & 92.4 \\
\hline
\end{tabular}
}
\label{tab:voxel_dim}
\end{table}

\subsection{Experiments with Different Voxel Dimensions}
Table~\ref{tab:voxel_dim} presents the results of our method with different bin sizes of the voxel grid. With the smallest bin size of 1, optimal performance is not achieved due to information loss, as much temporal information is aggregated into a single channel. However, as the bin size increases to 5, the performance improves because the discretely separated bins allow for the efficient utilization of temporal information. Nonetheless, when the bin size becomes excessively large, the events become spatially sparse, and the convolutional layers are unable to fully exploit this temporal information, resulting in a performance drop.


\subsection{Qualitative Ablation Study of Temporal Aggregation}
Fig.~\ref{fig:qual_dsec} shows comparisons between our temporal event stereo network and a single stereo network, where other components are kept constant while modules related to temporal aggregation, specifically feature warping and cost volume warping, are removed.
Temporal stereo, in contrast to single stereo, leverages preceding information continuously for compensation, enabling more accurate detail reconstruction of scenes. It also shows resilience in difficult conditions, including noisy night environments and instances of fewer events.

\setcounter{figure}{0}
\setcounter{table}{0}
\setcounter{algorithm}{0}
\setcounter{equation}{0}
\renewcommand{\thefigure}{B\arabic{figure}}
\renewcommand{\thetable}{B\arabic{table}}
\renewcommand{\thealgorithm}{B\arabic{algorithm}}
\renewcommand{\theequation}{B\arabic{equation}}
\renewcommand{\thesection}{B}

\begin{table}[p]
\centering
\caption{Structure details of the networks.}
\label{tab:Structure_Tables}
\renewcommand{\tabcolsep}{4.5pt}
\resizebox{0.95\textwidth}{!}{
\begin{tabular}{c|c|c}
\thickhline \textbf{Name} & \textbf{Layer setting} & \textbf{Output Dimension}\\
\thickhline Input & Voxel Grid & $H \times W \times 5$\\
\thickhline \multicolumn{3}{c} {\textbf{Feature Extractor}} \\
\thickhline Conv0 & $\left[3 \times 3, 12\right] \times 3$, stride=2 & $H / 2 \times W / 2 \times 12$ \\
\hline 
\rule{0pt}{2ex}
Conv1 & {$\left[\begin{array}{l}3 \times 3, 12 \\ 
3 \times 3,12\end{array}\right] \times 2$} & $H / 2 \times W / 2 \times 12$ \\ [2ex]
\hline
\rule{0pt}{2ex}
Conv2 & {$\left[\begin{array}{l}3 \times 3,24 \\
3 \times 3,24\end{array}\right] \times 3$}, stride=2 & $H / 4 \times W / 4 \times 24$ \\[2ex]
\hline
\rule{0pt}{2ex}
Conv3 & {$\left[\begin{array}{l}3 \times 3, 36 \\
3 \times 3, 36\end{array}\right] \times 2$} & $H / 4 \times W / 4 \times 36$ \\[2ex]
\hline
\rule{0pt}{2ex}
Conv4 & {$\left[\begin{array}{l}3 \times 3, 36 \\
3 \times 3, 36\end{array}\right] \times 2$, dilation=2} & $H / 4 \times W / 4 \times 36$ \\[2ex]
\hline
\rule{0pt}{2ex}
\multirow{3}{*}{Avg1} & $16 \times 16$ avg. pooling & \multirow{3}{*}{$H / 4 \times W / 4 \times 12$} \\
& $3 \times 3, 12$ & \\
& bilinear interpolate & \\
\hline
\rule{0pt}{2ex}
\multirow{3}{*}{Avg2} & $8 \times 8$ avg. pooling & \multirow{3}{*}{$H / 4 \times W / 4 \times 12$} \\
& $3 \times 3, 12$ & \\
& bilinear interpolate & \\ 
\hline
\rule{0pt}{2ex}
\multirow{3}{*}{Fusion} & Concat(Conv2, Conv4, Avg1, Avg2) & \multirow{3}{*}{$H / 4 \times W / 4 \times 12$} \\
& $3 \times 3, 36$ & \\
& $3 \times 3, 12$ & \\ 
\hline
\rule{0pt}{2ex}
\multirow{2}{*}{}Feature  & Spatial Warping, $\mathcal{W}_s$ (Eq.(1)) \& Concat & \multirow{2}{*}{$H / 4 \times W / 4 \times 12$} \\
Warping & $3 \times 3, 12$ & \\ 
\thickhline 
\multicolumn{3}{c} {\textbf{Stereoscopic Flow}} \\
\thickhline
\multirow{2}{*}{Flow0}  & Concatenate Left and Right & \multirow{2}{*}{$H / 4 \times W / 4 \times 12$} \\
 & $\left[3 \times 3, 24\right] \times 8$ & \\
\hline
Flow1  & Add Flow0 \& $3 \times 3, 3$ & $H / 4 \times W / 4 \times 4$ \\
  
\thickhline \multicolumn{3}{c} {\textbf{Initial Cost Volume}} \\
\thickhline
Cost Volume & Concatenate Left and Shifted Right & $D_{max} / 4 \times H / 4 \times W / 4 \times 24$  \\
\hline
3D-Conv0 & $\left[3 \times 3 \times 3, 12\right] \times 2$ & $D_{max} / 4 \times H / 4 \times W / 4 \times 12$  \\
\hline
3D-Conv1 & $\left[3 \times 3 \times 3, 12\right] \times 2$ & $D_{max} / 4 \times H / 4 \times W / 4 \times 12$  \\
\hline
3D-Conv2 &  Add 3D-Conv0 \& 3D-Conv1 & $D_{max} / 4 \times H / 4 \times W / 4 \times 12$ \\
\thickhline \multicolumn{3}{c} {\textbf{Initial Hourglass}} \\
\thickhline
3D-Stack0-0 & $7 \times 7 \times 7, 24$, stride=3 & $D_{max} / 12 \times H / 12 \times W / 12 \times 24$  \\
\hline
3D-Stack0-1 & $ 3 \times 3 \times 3, 24 $ & $D_{max} / 12 \times H / 12 \times W / 12 \times 24$  \\
\hline
3D-Stack0-2 & $\left[3 \times 3 \times 3, 24\right] \times 2$ & $D_{max} / 12 \times H / 12 \times W / 12 \times 24$  \\
\hline
3D-Stack0-3 & $3 \times 3 \times 3, 24$, add 3D-Stack0-1 & $D_{max} / 12 \times H / 12 \times W / 12 \times 24$  \\
\hline
\multirow{2}{*}{3D-Stack0-4} & deconv $7 \times 7 \times 7, 24$, stride=3 & \multirow{2}{*}{$D_{max} / 4 \times H / 4 \times W / 4 \times 12$} \\
& add 3D-Conv2 & \\
\hline
\multirow{2}{*}{3D-Output0} & $3 \times 3 \times 3, 12$ & \multirow{2}{*}{$D_{max} / 4 \times H / 4 \times W / 4 \times 1$} \\
& $3 \times 3 \times 3, 1$ & \\
\hline
Output0 & bilinear interpolate \& regression & $H \times W$\\
\hline
3D-Stack1-0 & $7 \times 7 \times 7, 24$, stride=3 & $D_{max} / 12 \times H / 12 \times W / 12 \times 24$  \\
\hline
3D-Stack1-1 & $ 3 \times 3 \times 3, 24$, add 3D-Stack0-3 & $D_{max} / 12 \times H / 12 \times W / 12 \times 24$  \\
\hline
3D-Stack1-2 & $\left[3 \times 3 \times 3, 24\right] \times 2$ & $D_{max} / 12 \times H / 12 \times W / 12 \times 24$  \\
\hline
3D-Stack1-3 & $3 \times 3 \times 3, 24$, add 3D-Stack0-1 & $D_{max} / 12 \times H / 12 \times W / 12 \times 24$  \\
\hline
\multirow{2}{*}{3D-Stack1-4} & deconv $7 \times 7 \times 7, 24$, stride=3 & \multirow{2}{*}{$D_{max} / 4 \times H / 4 \times W / 4 \times 12$} \\
& add 3D-Conv2 & \\
\hline
\multirow{2}{*}{3D-Output1} & $3 \times 3 \times 3, 12$ & \multirow{2}{*}{$D_{max} / 4 \times H / 4 \times W / 4 \times 1$} \\
& $3 \times 3 \times 3, 1$ & \\
\hline
Output1 & bilinear interpolate \& regression & $H \times W$\\
\thickhline \multicolumn{3}{c} {\textbf{Cost Volume Refinement \& Aggregation}} \\
\thickhline 
\multirow{2}{*}{}Warped & \multirow{2}{*}{3D Warping, $\mathcal{W}_c$ (Eq.(3)) to Prev Final Cost} &\multirow{2}{*}{$D_{max} / 4 \times H / 4 \times W / 4 \times 12$} \\
 Cost Volume  &  \\
\hline
Entropy Filter & Generate the entropy filter based on 3D-Output1 & $ H / 4 \times W / 4 \times 1$  \\
\hline
\multirow{3}{*}{Weight0 \& Weight1} & Concatenate the entropy filters & \multirow{3}{*}{$ H / 4 \times W / 4 \times 2$}  \\
&  $\left[3 \times 3 , 12\right] \times 3$\\
& $3 \times 3 , 2$ \& Sigmoid &\\
\hline
Fused Cost Volume & Weight0 $\times$ 3D-Stack1-4 + Weight1 $\times$ Warped Cost & $D_{max} / 4 \times H / 4 \times W / 4 \times 12$  \\
\thickhline \multicolumn{3}{c} {\textbf{Final Output}} \\
\thickhline
3D-Stack2-0 & $7 \times 7 \times 7, 24$, stride=3 & $D_{max} / 12 \times H / 12 \times W / 12 \times 24$  \\
\hline
3D-Stack2-1 & $ \left[3 \times 3 \times 3, 24\right] \times 2 $  & $D_{max} / 12 \times H / 12 \times W / 12 \times 24$  \\
\hline
3D-Stack2-2 & $ \left[3 \times 3 \times 3, 24\right] \times 3$, add 3D-Stack3-1  & $D_{max} / 12 \times H / 12 \times W / 12 \times 24$  \\
\hline
\multirow{2}{*}{3D-Stack2-3} & deconv $7 \times 7 \times 7, 24$, stride=3 & \multirow{2}{*}{$D_{max} / 4 \times H / 4 \times W / 4 \times 12$} \\
& add 3D-Conv2 & \\
\hline
\multirow{2}{*}{3D-Output2} & $3 \times 3 \times 3, 12$ & \multirow{2}{*}{$D_{max} / 4 \times H / 4 \times W / 4 \times 1$} \\
& $3 \times 3 \times 3, 1$ & \\
\hline
Output2 & bilinear interpolate \& regression & $H \times W$\\
\hline
\end{tabular}}
\end{table}

\setcounter{figure}{0}
\setcounter{table}{0}
\setcounter{algorithm}{0}
\setcounter{equation}{0}
\renewcommand{\thefigure}{C\arabic{figure}}
\renewcommand{\thetable}{C\arabic{table}}
\renewcommand{\thealgorithm}{C\arabic{algorithm}}
\renewcommand{\theequation}{C\arabic{equation}}
\renewcommand{\thesection}{C}

\begin{figure*}[t]
\begin{center}
\includegraphics[width=1\linewidth]{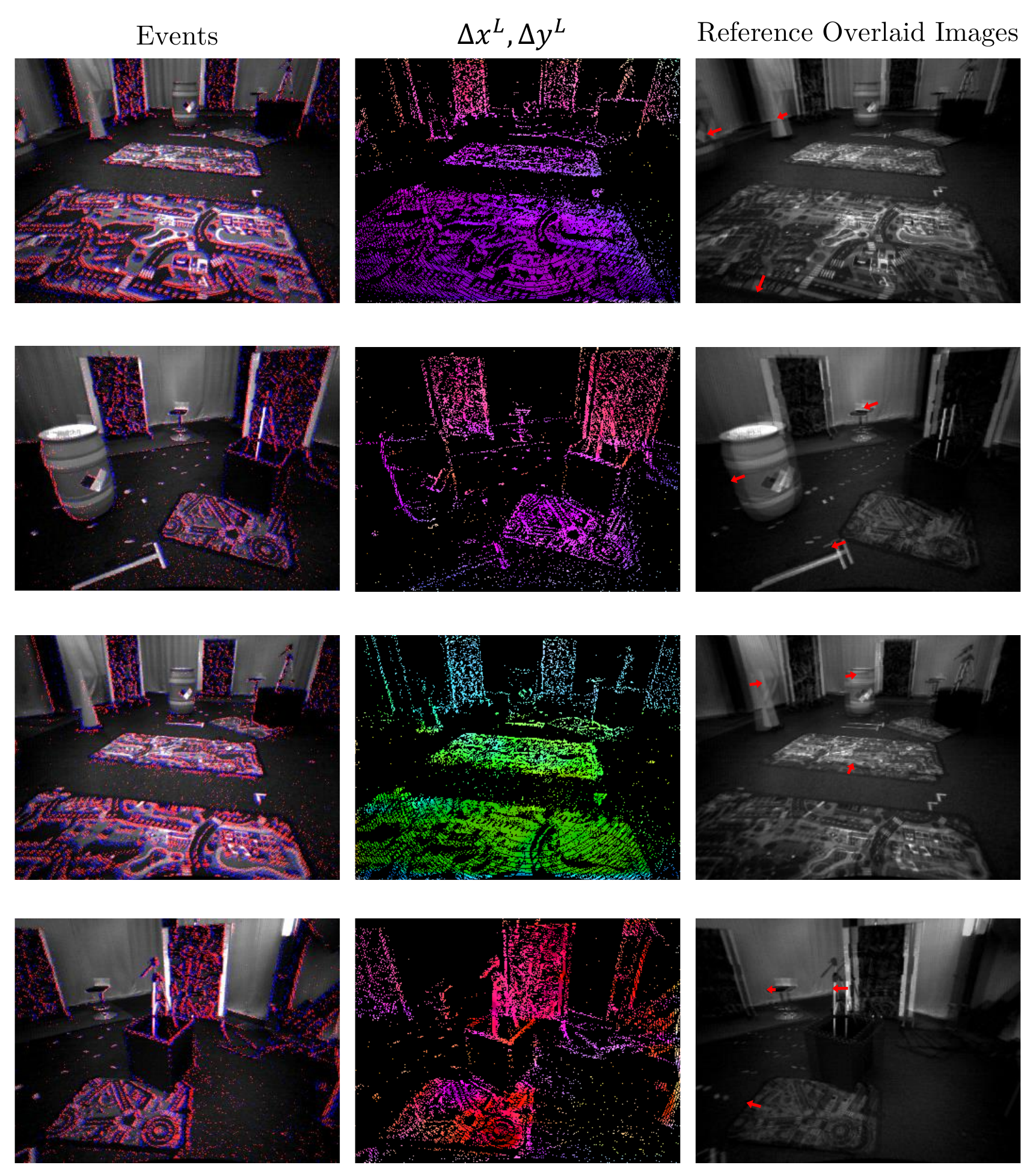}
\caption{Visualization of the components $\Delta x^L$ and $\Delta y^L$ of the left camera in stereoscopic flow. For the visualization, we set the color wheel identical to that of~\cite{baker2011database}. To make the direction of motion clear, we presented a visualization by overlaying the current image with the one from three frames earlier, indicating the direction of forward movement using \textcolor{red}{red arrows}. \textbf{Note that our stereoscopic flow is a backward flow}.}
\label{fig:qual_flow_mvsec}
\end{center}
\end{figure*}

\begin{figure*}[t]
\begin{center}
\includegraphics[width=1\linewidth]{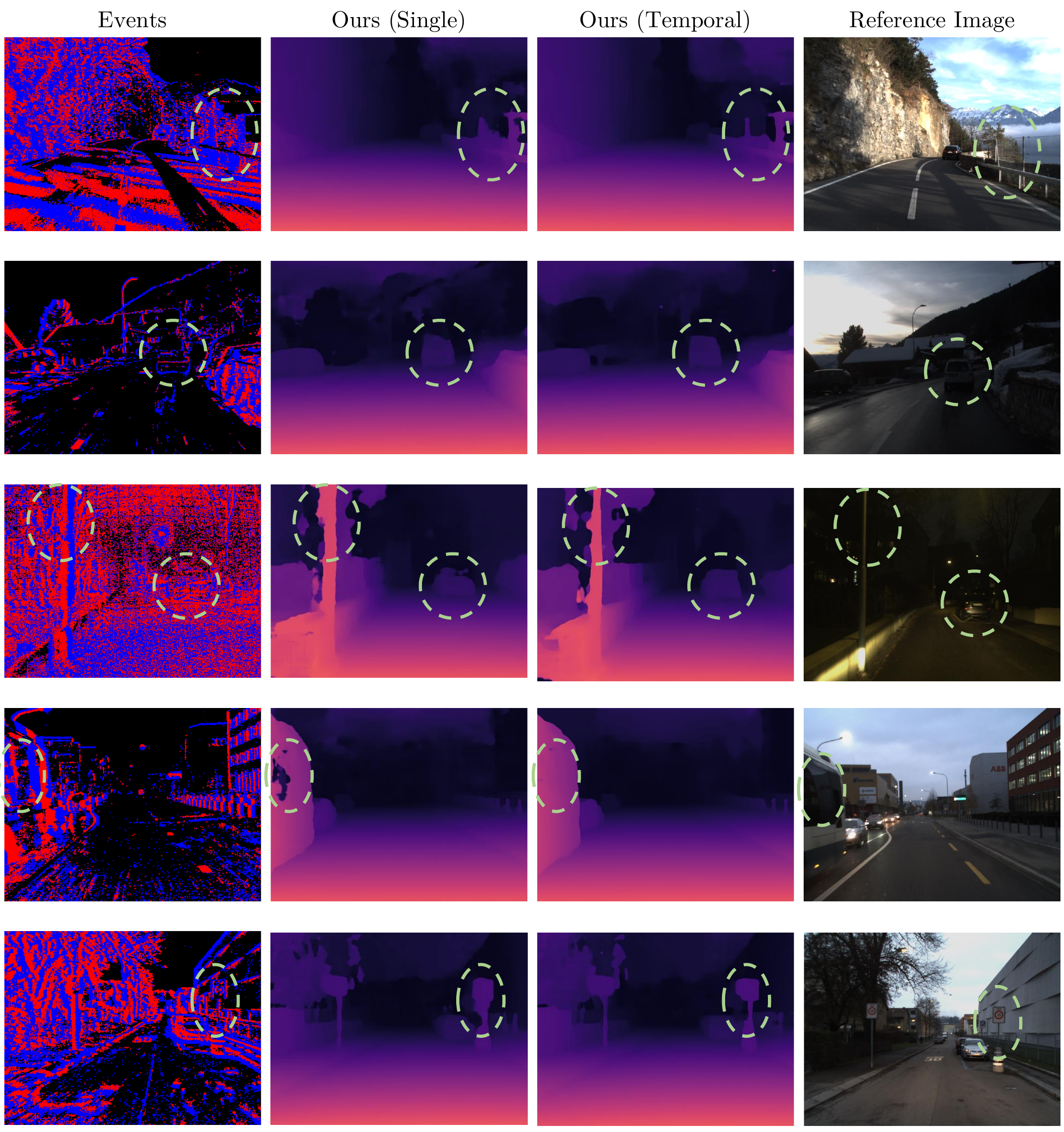}
\caption{Qualitative comparison between single and temporal event stereos on DSEC test datasets.}
\label{fig:qual_dsec}
\end{center}
\end{figure*}




\clearpage

%
%
\end{document}